# A Review of Published Machine Learning Natural Language Processing Applications for Protocolling Radiology Imaging

Running title: A Review of AI in Radiology Protocolling

Nihal Raju[*,5], Michael Woodburn[*,1,5], Stefan Kachel[2,3], Jack O'Shaughnessy[5], Laurence Sorace[5], Natalie Yang[2], Ruth P Lim[2,4]

Austin Health, 145 Studley Rd, Heidelberg, VIC, 3084, Australia

Corresponding author: Michael Woodburn

https://orcid.org/0000-0003-0600-8866

woodburnmichael@gmail.com

Austin Health, 145 Studley Rd, Heidelberg VIC 3084, Australia

Emails:

NR: nihal.r.raju@gmail.com

MW: woodburnmichael@gmail.com

SK: Stefan.KACHEL@austin.org.au

JO: jack.o'shaughnessy@austin.org.au

LS: laurence.sorace@bigpond.com

NY: Natalie.YANG@austin.org.au

RL: Ruth.LIM@austin.org.au

---

[*] These authors contributed equally
[1] Harvard University, Extension School, Cambridge, MA, USA
[2] Department of Radiology, The University of Melbourne, Parkville
[3] Department of Radiology, Columbia University in the City of New York
[4] Department of Surgery (Austin), The University of Melbourne
[5] Austin Hospital, Austin Health, Melbourne, Australia

# Abstract and Keywords

## Abstract

Machine learning (ML) is a subfield of Artificial intelligence (AI), and its applications in radiology are growing at an ever-accelerating rate. The most studied ML application is the automated interpretation of images. However, natural language processing (NLP), which can be combined with ML for text interpretation tasks, also has many potential applications in radiology. One such application is automation of radiology protocolling, which involves interpreting a clinical radiology referral and selecting the appropriate imaging technique. It is an essential task which ensures that the correct imaging is performed. However, the time that a radiologist must dedicate to protocolling could otherwise be spent reporting, communicating with referrers, or teaching. To date, there have been few publications in which ML models were developed that use clinical text to automate protocol selection. This article reviews the existing literature in this field. A systematic assessment of the published models is performed with reference to best practices suggested by machine learning convention. Progress towards implementing automated protocolling in a clinical setting is discussed.

Key Words*:*

Artificial Intelligence, Machine Learning, Natural Language Processing, Radiology, Review Literature as Topic

# Background

Machine learning (ML) is a subfield of Artificial Intelligence (AI) that describes computer algorithms which learn relationships and patterns within data in order to produce a desired output(1). ML has increasingly been applied in Radiology to interpret images. It is commonly used to automate tasks where radiologists interpret features within images to reach medically relevant conclusions. A recent systematic review showed that deep learning, a subtype of ML which uses artificial neural networks(2), had equivalent performance to healthcare professionals in detecting some diseases in medical images (3). ML can be used for other applications beyond diagnosis. For example, a recent study described a successful ML model that identified biomarkers in body CT scans for cardiovascular risk stratification which were more predictive than clinical parameters (4). ML has also been used to improve radiologist efficiency, such as by automating the triaging of adult chest radiographs, which significantly reduced reporting delays for images with critical findings. (5)

Natural language processing (NLP) is another subfield of AI that describes techniques and processes which "convert unstructured text into a structured form, and therefore enable automatic identification and extraction of information" (5). NLP has not been as widely studied for Radiology applications, however, since radiologists must document their findings, interpretation, and recommendations in written reports, there are many theoretical applications which make use of that text data. Examples of target outcomes include automated identification of follow up recommendations (6), automated extraction of the diagnosis from a report (7), or classification of report free text into discrete labels.(8)

# Background

Another example of free text data in Radiology is the clinical referral for a scan. Study protocolling for computed tomography (CT) and magnetic resonance imaging (MRI) scans relies upon clinical text information provided by the referrer to generate a written 'prescription' by a radiologist for the optimal way to perform the requested study to answer the clinical question (Fig 1).

A correctly prescribed protocol can improve patient outcomes and reduce costs (9, 10). An incorrect, inappropriate, or incomplete protocol can result in wasted resources, delayed diagnosis, or unnecessary exposure to ionising radiation in CT studies (10). One study calculated that protocolling occupies up to 6.3% of a radiologist's time (9), therefore its automation would conceivably allow more time for image interpretation, interventional procedures and direct communication with referrers and patients.

The following is an example of a referral and corresponding protocol from one of the included studies(11):

Referral:

*Indication: long smoking history and weight loss. rule out malignancy*

Protocol:

*CT Chest without contrast*

In this review article, we begin with a brief overview of ML theory. We then present an in-depth systematic review of the existing literature for NLP for Radiology protocolling, with reference to relevant elements of the recently published Checklist for Artificial Intelligence in Medical Imaging (CLAIM) guidelines, which set a standard that Radiology studies should meet to legitimately report the development of a successful ML model(12). Finally, we summarise progress to date and discuss future directions to bring automated protocolling into the clinical domain.

## Methods: Search Strategy

Embase and Medline (Ovid) online databases were searched in February 2021 using keywords and medical subject headings. Results were limited to English. Articles were incorporated if they demonstrated the development of an automated protocolling model related to any imaging modality according to PRISMA guidelines(13). The search strategy and selection criteria are summarized in Appendix i and ii.

Based on our search strategy, five studies of radiology protocolling using ML were identified (Table 1).

## Machine Learning:  A Brief Overview

*Figure 1*

Developing a ML model for protocolling is an example of supervised ML (Figure 2). In supervised ML, during training, the model is shown the ground truth labels. For the developed models which this review assesses, we discuss the selected task, data, hardware, software, and evaluation strategy.

ML automation is generally strongest for tasks similar to protocolling scans, i.e. tasks that are repetitive and have a clearly defined and valuable objective. To ensure that the proposed task has these qualities radiologist in-domain expertise is essential.  Experience as a machine learning practitioner is also often necessary to confirm that automation of a task is technologically feasible.

Supervised ML requires a dataset of pairs of inputs and associated labels (radiologists may refer to these as reference standards, or ground truths). A label is an indicator that a sample belongs to a specific class. For protocolling, each label is a specific protocol selected by an expert as the most appropriate protocol for a given text referral. Labelling by clinical experts is recommended by most guidelines, although it often limits the size of the labelled dataset due to labour costs.

The tools used in ML include hardware capable of storing data, a Central Processing Unit (CPU) or Graphics Processing Unit (GPU) for efficient calculations, and software to develop, train and test an appropriate model. Since NLP models use text as inputs, and text inputs are often processed by deep learning models, not all models can be trained on ordinary computers.

Evaluation of a model uses metrics (e.g. recall and precision) for the proposed task, and can include assessment of the model's generalisability to previously unseen data. Evaluation must be done according to specific conventions to be considered statistically valid.

## The Task: Protocol Selection

Radiology protocolling lends itself well to ML, since it is a task that is completed frequently by radiologists and results in an abundance of data already stored in electronic format in Radiology Information Systems (RIS). Radiology referrals with clinical indications (inputs) and radiologist-informed protocols (labels) can be retrospectively extracted and used for model training and testing.

All identified studies used retrospectively sampled MRI scan data, except for Kalra et al. (11) which also included CT data. Tasks included prescribing each specific sequence to be performed during an examination (14), determining whether contrast administration is necessary and, in Brown et al. (15), determining whether a radiological scan should be prioritised to occur within 24 hours.

The proposed benefits of automated protocolling highlighted by the studies were: achieving appropriate and consistent selection of radiological examination (14), improved efficiency (11, 16, 17) and prioritisation of urgent examinations(15), with downstream benefits to patient care, safety (mitigating inappropriate ionising radiation or contrast) and a reduction in overall health costs (17).

*Figure 3:*

## Data

Good quality data is an integral determinant for the success of a ML model. Radiologists will be familiar with the old adage, "garbage in, garbage out" as applied to radiology requests, also recognised in data science as the acronym *GIGO* (18). Thus, quality data is essential for developing an automated protocolling model. Referral quality is determined by the referring doctor writing the clinical indication, and protocol selection quality based on the referral is the responsibility of the radiologist.

## Referrals

Referring physicians are directed to include necessary information to choose a protocol to match the clinical indication. If the appropriate protocol is not clear for a radiologist from the provided indication, it will likely be similarly unclear to a model. Four out of five studies used unadjusted clinical text as inputs into their data pipelines (some included additional inputs), and each of their models had to contend with this inherent uncertainty(11, 14, 15, 17). For example, a referral of 'abdominal pain' could lead to several protocols. If calculus is suspected, a non-contrast CT KUB would be selected. If appendicitis is favoured, a portal venous phase CT would be more appropriate. A suspicion of mesenteric ischaemia would require multi-phase imaging. Referrer communication and education has been historically emphasized as an area of improvement for optimising radiology examination appropriateness (19, 20).

To potentially address this issue, information beyond the referral can be gathered from the electronic health records (EHR) as ancillary input data. This could include past medical history and estimated glomerular filtration rate, which have been recommended as inputs to improve the appropriateness of the output protocol (11, 14, 17). Trivedi et al. and Brown et al. explicitly used human-authored clinical indication as the sole input (14, 17). In contrast, Kalra et al. included multiple fields in its inputs including clinical history, some of which was auto-generated by the Computerized Patient Record System (11). Lee et al.'s inputs were a combination of the referring department, the region of interest, whether contrast was used, gender, and age of the patient (16).

In all five studies assessed, only single institution data was used (11, 14, 17). Kalra et al. collected the largest dataset (18,192 samples) (11) and Trivedi et al. the smallest dataset (1,520 samples), perhaps reflecting the latter's choice to manually review and classify each sample (17). The more data available for training in ML, the better the chance the algorithm has of learning relevant features and dealing with previously unseen data. A notable example is the RSNA dataset of intracranial haemorrhage which contained 25,000 CT brain images and resulted in the development of highly accurate models (21).

## Protocol labels

Of equal importance to the input data is the quality of its associated output – the label. Labels provide the ground truth which the model uses to learn. Ideally, a clinical referral has a one-to-one relationship with its protocol i.e., the same clinical referral is protocolled the same way consistently. However, for many reasons, one-to-many relationships between clinical referrals and protocol labels stored in RIS exist, which disrupts training in ML(15). Variation in protocols may be caused by differences in the level of experience of the individual choosing the protocol (e.g. radiographer versus radiology trainee versus experienced radiologist), or intra-individual variation in protocols due to human error.

For reliable results, accurate labelling of data should be performed by at least one expert radiologist for radiology protocols (22). Consensus labelling with more than one radiologist performing labelling is even more reliable than a single expert radiologist. Of the five studies, only Kalra et al. and Trivedi et al. stated that the protocol label had been confirmed by a fully trained radiologist, as opposed to a trainee radiologist or a radiographer (17), which strengthens the validity of their test metrics.

In clinical practice, the protocol label to be assigned to a referral is usually selected from a predefined set during clinical workflow for ease of implementation and efficiency (Figure 3). Labelling from a predefined set is an advantage to ML classification models, because it limits the number of possible outputs. In general, for classification tasks, fewer potential label classes result in better performance scores on common metrics, since the probability of a correct classification is not 'diluted' across numerous label classes.

# Hardware and Software

*Table 2*

ML involves computations which improve the model during training so that its outputs become more accurate to the ground truth labels. In assessing the computational hardware and software, we ascribe high value to methods that adhere to ML best practices, as well as those that are replicable by other researchers.

## Hardware

Classic ML models (e.g Support Vector Machine, Random Forest, Logistic Regression) as used in four of the studies(11, 14, 15, 17), typically have hardware requirements that can be met by standard personal computers equipped with a CPU, thus posing no barrier to entry for the typical researcher. In contrast, the training of deep learning models on large datasets usually requires access to a quality GPU, which decreases training time by orders of magnitude(23).

Three studies used deep learning methods. Lee et al. employed a commercially available GPU(16). Kalra et al. atypically trained a simple deep learning model without a GPU(11), which is sometimes feasible due to small dataset size. Trivedi et al. used the IBM Watson service, which makes use of server-side hardware remote from the user (17). Such cloud-based services offer, by a large degree, the greatest computing power the typical programmer can access.

## Software

In both studies by Brown et al. and Trivedi et al. (14, 15, 17), classic ML models were trained in the R programming language. In Kalra and Lee (11, 16) the language employed was Python, a competitor to R. The two languages have some overlap in their domains; Python is more popular and more versatile, and R is statistics and machine-learning specific(24). Whilst in 2017 the growth of the two languages was equal (25), in the last three years Python has outpaced R and is now the dominant language in machine learning(26). Neither language will, *ipso facto*, develop superior ML models, but due to the popularity of Python, the models developed with it will likely be more accessible and reproducible for other scientists.

In addition to the language, the 'libraries' for training can significantly affect reported results. By using popular open-source ML libraries in Python, sci-kit learn (27) and Keras (28), Kalra et al. (11) increases the ease with which those methods can be implemented by other users. In contrast, Trivedi employed IBM Watson, a paid ML service (17), with its operations obscured from the user. It is difficult to achieve deterministic results and to replicate experiments with such a service. However, its availability as a cloud service means that experiments can be implemented quickly by researchers who pay for the service. Whilst cloud-based ML solutions likely represent the future of research, they also create potential for privacy and data security breaches and may require additional security measures to protect patient data (29, 30).

## Model Training

### Representation and pre-processing

A supervised ML NLP model receives text data created by humans as an input. In computer science, this input is known as free text. Free text must be transformed into a computer-readable format, for example, by representing the presence of absence of a word with a 1 or 0. The diversity of written language creates a great number of variables, generating 'noise' in the data which might obscure the real relationship between inputs and outputs (31). Pre-processing text is a strategy to represent text without losing meaning and reduce noise before it is presented as an input to the model.

In all included studies, two common pre-processing methods were employed. 'Stopwords', which are words so common in the English language that they convey little meaning (e.g. 'the'), were removed. *Lemmatisation* was also employed, in which words are transformed into their dictionary form by removing inflectional endings. (e.g 'scanned', 'scans', and 'scanning' all become 'scan').

Neural network (deep learning) models are commonly combined with a type of pre-processing termed embedding. In *embedding,* words with similar meanings are presented to the model as 'close together' in vector space, allowing for a memory-efficient and meaning-rich representation of text. In Lee et al. (16), word embedding was used. Conversely, although Kalra et al (11) employed a neural network, embeddings were not used, thus demonstrating that although effective, they remain an optional pre-processing step in deep learning (32).

To achieve the highest model performance, an optimal method of pre-processing is usually selected, which is usually determined by adjusting pre-processing parameters iteratively (33). It is not stated whether multiple pre-processing methods were trialled in the five studies reviewed, and it is not discoverable since development code has not been made publicly available.

## Model selection

The model is the algorithm which learns, for protocolling purposes, which protocol a radiology referral should be assigned. Models are divided into two classes: classical ML and deep learning models.

Classical ML models include k-Nearest Neighbours, which predicts that samples, when represented as points in space, are likely to be of the same class (for our purposes the class is the protocol) as samples which are close to them. Support Vector Machines represent another important classical ML technique, which works similarly but instead creates a boundary around points of the same class. Points within that boundary are predicted to belong to that class. Classical models were used in four studies (11, 14, 15, 17).

Deep learning models, which can learn higher level abstractions of input features, such as those used in Kalra, Trivedi, and Lee (11, 16, 17), tend towards better performance as the dataset size increases(34). This was the case in Kalra (11) where the deep neural network slightly outperformed the random forest (classical ML) approach, with a dataset of large size (18,192 samples).

Best practice in model selection involves experimenting with many models and maintaining an open mind as to which will achieve the highest performance. Reassuringly, all papers except Lee et al. experimented with multiple models (Table 2) and compared their results (16). Lee only used a CNN model, and a comparator classical model was not developed on their dataset (16).

# Evaluation and Optimisation

*Table 3:*

Every task has specific criteria for success. However, in ML, there are conventional ways of proving that a model works by using 'metrics', which can be thought of as performance scores. Models are scored on how they perform on internal data, which is usually sourced from the same institution(s) as the training data, and models can also be scored on how they perform on external data, which is often sourced separately, preferably from another institution.

## Internal Data

In ML, the dataset of referral-protocol pairs is split into at least two sets, *training,* and *test* (Figure 3). A *validation* set is sometimes split from the *training* set (typically approximately 10-20% of the training set). The *training* and *validation* sets are used to optimise the model. Models learn from the training set, are fine-tuned on the validation set, and their accuracy is reported on the test set.

The way these splits are chosen is vital to the validity of the study, as a contrived division of samples can be used to artificially increase the performance of the model and exaggerate its quality (35). Therefore, the splits should be made according to a standard set of rules. Commonly, approximately 80% of the entire sample goes into the training and validation sets; and 20% are held in the test set (36).

In all five studies, this convention was abided by with minor variations. Kalra (11) reported an 80-20 split *by protocol*, to achieve proportional representation of all classes in the test set. Trivedi (17), similarly, opted for an 82-18 training/validation-test split, albeit also implementing weighted sampling so that the test set contained an equal amount of both classes (not proportionate to their actual incidence, and therefore likely to result in a different reported accuracy than if the model were tested on representative data). In Lee (16), all the protocols in the test set were selected from a one-month period, January to February 2017, but the justification for choosing that time period was not provided. The variability in dataset division methods suggests a lack of accepted convention regarding data partition in Radiology ML at the time of the studies' publication, now addressed by CLAIM (12).

### Test set integrity

The performance of the model on the test set was reported in all five NLP protocol studies (11, 14-17) to grade performance. The integrity of the test set is a fundamental requirement to ensure an accurate reflection of model performance. This has its pitfalls, as the accuracy of the model on the test set can be artificially elevated by at least two common means.

First, when the test set is visible to the researchers, the model can be trained with many random adjustments to its internal variables known as 'hyperparameters' over a massive number of iterations until a model is achieved which achieves an artificially high performance on the test set. Secondly, a model can be trained with constant hyperparameters but the dataset can be split in many random ways until, by pure chance, a test set is generated on which the model achieves an artificially high performance.

The integrity of the results can only be upheld if the test set is withheld from and invisible to the researchers until they have finalized their model. This requires a setup of considerable sophistication and can be seen in practice in radiological data science competitions (37-39). None of the experiments reviewed in this paper provided clear indication that the integrity of the test set was maintained at this standard, although this detail may not have been explicitly stated.

## Performance metrics

Interpreting metrics

When interpreting metrics, consideration should be made towards the priority of recall (identifying as many of the true positives as possible) versus high precision (having confidence that a predicted positive result is a true positive). There is usually a trade-off between these two metrics. In medicine, high recall reduces missed diagnoses. All five studies reported recall and precision or a variation thereof.

The two studies which developed binary classification models (two possible labels) used similar metrics. In Lee et al. (16), the metrics Area Under the Receiver Operating Curve (AUROC) (40), sensitivity, specificity, positive predictive value (PPV), negative predictive value (NPV) are appropriately used since classes are approximately balanced (and binary) at 2043 tumour and 3215 routine protocols respectively. Trivedi used the same, except for AUROC. In Brown (2017) (15), similarly, metrics are used appropriately for the two binary classification subtasks where both classes were relatively balanced.

For multiclass classification, metrics for each *individual* class are required to validate a model's performance. However, in general, for the studies which developed multiclass classification models, Brown (2017), Brown (2018), and Kalra (11, 14, 15), model performance was not reported for each individual class. Only the mean metrics for all protocols were presented. This may conceal poor performance for uncommon protocols. This was particularly relevant for Kalra et al. (11), with 108 possible protocols and significant class imbalance. The use of mean metrics, and especially 'accuracy', is insufficient to report the specific weaknesses of trained models, with greater rigour encouraged in the CLAIM guidelines (12).

However, perhaps to combat the large number of possible protocol outcomes, Kalra (11) used an innovative approach to metric selection. It was proposed that where the model's own confidence in correctly selecting a single protocol was low, the model could output the *three* most likely protocols. Returning such a result was termed 'Clinical Decision Support' and could conceivably be useful to aid an inexperienced individual with protocolling. The model achieved a high accuracy according to this proposed metric (91.5%). The validity of applying a model in this way should be peer reviewed by domain-specific experts, in this case, radiologists.

## Explaining how the model works

In ML, the 'black box' problem exists, whereby it may not be possible to meaningfully explain how a model produces certain outputs(41). The CLAIM guidelines (12) suggest that all AI medical imaging studies should include an attempt to interpret how a model works. Saliency graphs achieve this by showing what features in an image or words in a string of text a model interprets as 'important'. Even though many of the papers reviewed here did in fact use interpretable models, only Brown et al. (15) provided a list of words ordered by their importance to the model. In the other models, the features deemed important to the model were not reported.

Another way to explain a model is error analysis, which is an oft-cited requirement for the progression of medical AI (42). Kalra et al. evaluated the model's errors by comparing the performance of two expert radiologists with each other and the model for protocolling (11).

These experts were shown to be only 36% concordant on referrals where the model was in error, suggesting that the model was failing at protocolling ambiguous clinical referrals that also challenged domain experts. Trivedi et al. (17) listed examples of referrals for which the model produced false negative results and offered hypotheses for the errors.

## External Validation

Trained models should also be tested on datasets sourced externally to that used in their development (22). If a model can perform well on external datasets, it is likely to generalise well to other institutions, which is desirable for widespread or commercial use.

None of the ML protocolling studies to date performed external validation, with models all trained from data sourced from a single institution and tested on internal data from the same institution. Therefore, the generalisability of the developed models remains unproven. To our knowledge, there is presently no publicly available dataset of referral-protocol pairs which could facilitate external validation.

### Model Availability

Although not a requirement of CLAIM, there are benefits to a trained NLP model being made publicly available. Its performance can be verified by independent groups with external datasets (22), or used for further model development in a process known as transfer learning (43). None of the studies reviewed made their code or trained model publicly available.

### Regulatory Approval

The regulatory landscape for AI models in medicine is complex and under development. Notably, 64 AI algorithms, 30 of which are imaging-based as of 2020, have already been cleared for commercial use in the United States (44, 45). In general, models that play a clinical support role rather than fully independent role carry lower risk and, to date, have been more readily approved. The Royal Australian and New Zealand College of Radiologists recently released a Standards of Practice for Artificial Intelligence document to guide ethical use of AI in the field (46).

The five studies reviewed stated hypotheses were to, in general, assess the accuracy of a model for automated protocolling. For regulatory approval to be given, a study would also need to be designed to fulfil the requirements of the relevant regulatory body, which typically demand more extensive evaluation than that published in the studies reviewed.

## Conclusion

We have assessed the existing literature for automated protocolling of radiology examinations using ML. Overall, the studies demonstrated feasibility, but only partially fulfilled expert guidelines necessary for rigorous validation. Whilst internal validity was evaluated, external validity and clinical utility were not formally assessed or reported. Future directions include sharing the models and accompanying code from these studies, enabling others to verify and potentially improve model performance. Sharing the (deidentified) datasets would facilitate more rapid development of more accurate algorithms, as has been observed in other domains with datasets including Word-Net for language models (47) and ImageNet for image-based tasks (48), or for Radiology, with the CheXpert and MIMIC-CXR datasets for chest X-ray interpretation (49, 50). Application of more powerful recently developed NLP models, such as Bidirectional Encoder Representations from Transformers (BERT)(51), could further improve performance. Deep learning hardware advances, cloud computing services such as Google Colab (52), and user-friendly software are readily available for application. With scientific collaboration, current technological assets, clear domain-specific guidelines (12) and appropriate regulatory approval, automated protocolling can be translated in the near future to the clinical domain.

Tables

*Table 1: The five included studies in this review and the selected task(s) for each developed model*

| First Author and Year | Protocol Task | Task type | Possible Classes | Balanced Classes | Value proposition |
|---|---|---|---|---|---|
| *Kalra A et.al (2020)(11)* | Select the protocol of multisystem CT and MRI scans | Multiclass classification | 108 unique standard protocols | No | Enhance multispeciality CT and MRI protocol assignment quality and efficiency. |
| *Trivedi H et.al (2018)(17)* | To determine whether to use intravenous contrast protocol for musculoskeletal MRI protocols. | Binary classification | 2 protocols: NC (Non-contrast), WC (With contrast) | Yes | Improve efficiency and one day serve as a decision support tool for contrast assignment. |
| *Lee YH (2018)(16)* | Select the musculoskeletal MRI scan protocol from two options | Binary classification | 2 protocols: Tumour/Infection and Routine | Yes | Provide timely and highly accurate protocol determinations that only require rapid confirmation. |
| *Brown AD et.al (2018)(14)* | Select the protocol for MRI brain studies. | Multiclass classification | 41 unique standard protocols. | No | To guide sequence acquisition decisions and potentially improve efficiency, quality, and cost. |
| *Brown AD et. al (2017)(15)* | Select the protocol of MRI brain examinations from MRI requisitions | 1. Multiclass classification | 13 unique standard protocols. | No | To streamline the patient experience and improve the safety and efficiency of imaging procedures. |
| | Select whether contrast is given in MRI brain examinations | 2. Binary classification | 2 protocols: with contrast or without. | Yes | |
| | Select the priority of MRI brain examinations | 3. Binary classification | 2 protocols: Study performed within 24 hours or not. | No | |

*Table 2: Hardware and software (including specific ML model algorithms) used in the included studies.*

*† SVM: Support Vector Machine, SLDA: Scaled Linear Discriminant Analysis, CART: Classification and Regression Tree, GLMNET: Lasso and elastic-net regularized generalized linear model*

*‡ DNN: Deep neural network, CNN: Convolutional neural network.*

*§ GPU: Graphics Processing Unit.*

| First Author and Year | Cloud or Local Machine | Language | Libraries | Preprocessing | Classic Models† | Deep Learning‡ | GPU§ |
|---|---|---|---|---|---|---|---|
| *Kalra A et.al (2020)(11)* | Local | Python | Scikit-Learn, Keras | Lemmatization, Stopwords removed, TF-IDF | k-Nearest Neighbours, Random Forest | DNN | No |
| *Trivedi H et.al (2018)(17)* | Local and Cloud | R | R NLP libraries, RTextTools | Lemmatization, Stopwords removed, | SVM, SLDA, boosting, Bagging, CART, Random Forest, GLMNET, maximum entropy | Watson | Yes |
| *Lee YH (2018)(16)* | Local | Python | Tensorflow, Keras | Lemmatization, Stopwords removed, embedding | None | CNN | Yes |
| *Brown AD et.al (2018)(14)* | Unknown | R | R libraries | Lemmatization, Custom stopwords removed | SVM, Gradient Boosting Machine, Random Forest | No | Unknown |
| *Brown AD et. al (2017)(15)* | Unknown | R | R libraries | Lemmatization, Stopwords removed, acronyms expanded | Random Forest, SVM, k-neareset neighbour | No | Unknown |

*Table 3: Comparison of datasets obtained, and evaluation strategies incorporated*

† *PPV: Positive Predictive Value, NPV: Negative Predictive Value, AUC: Area Under the Receiver Operating Curve.*

| First Author and Year | Dataset size | Train : test dataset split | Split process | Metrics Used† | Key result emphasized in discussion | Model behaviour explained or interpreted | Code and model weights available online | Tested on external data |
|---|---|---|---|---|---|---|---|---|
| *Kalra A et.al (2020)* | 18192 | 0.8 : 0.2 | Random at protocol level | Weighted Precision and Recall | 95% accuracy in automation mode. 92% accuracy in clinical decision support mode | Yes, errors interpreted | No | No |
| *Trivedi H et.al (2018)* | 1520 | 0.82 : 0.18 | Random with weighted sampling to achieve 50-50 balance. | Sensitivity, Specificty, PPV, NPV, Accuracy | 90% accuracy | Yes, errors interpreted | No | No |
| *Lee YH (2018)* | 6276 | 0.84: 0.14 | Subsequent in time | AUC, Sensitivity, Specificity, PPV, NPV, Accuracy | 94% accuracy | No | No | No |
| *Brown AD et.al (2018)* | 7487 | 0.7 : 0.3 | Random | Accuracy, Precision, Recall, Hamming Loss | 95% accuracy | No | No | No |
| *Brown AD et. al (2017)* | 13,982 | 0.8 : 0.2 | Random | Accuracy, Sensitivity, Specificity, PPV, NPV, AUC | 83% accuracy | Word importance scored | No | No |
| | 10,100 | | | | 83% accuracy | | | |
| | 10,100 | | | | 88% accuracy | | | |

# Figures

*Figure 1: The steps in assigning a protocol to a referral in order to achieve a scan*

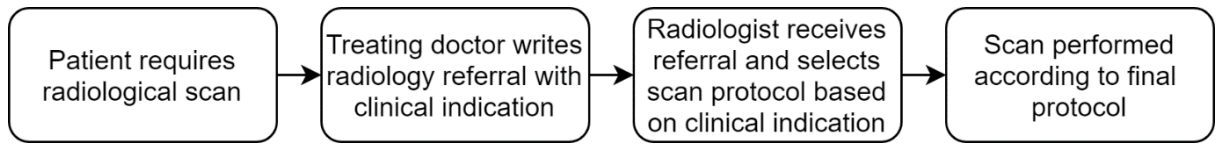

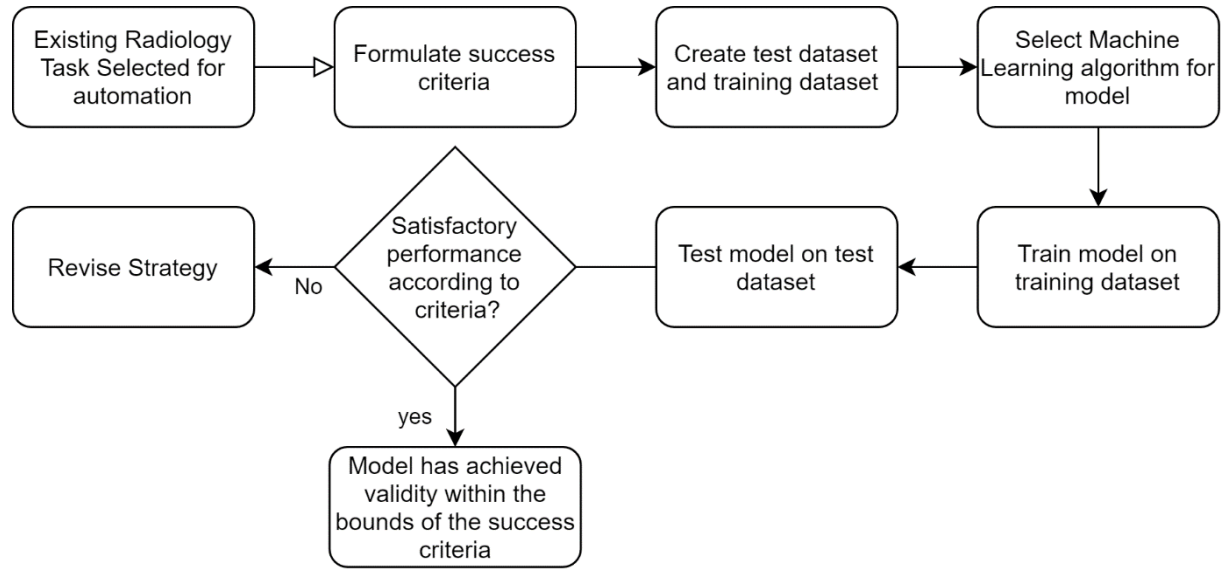

*Figure 2: Steps in developing and testing a supervised ML model*

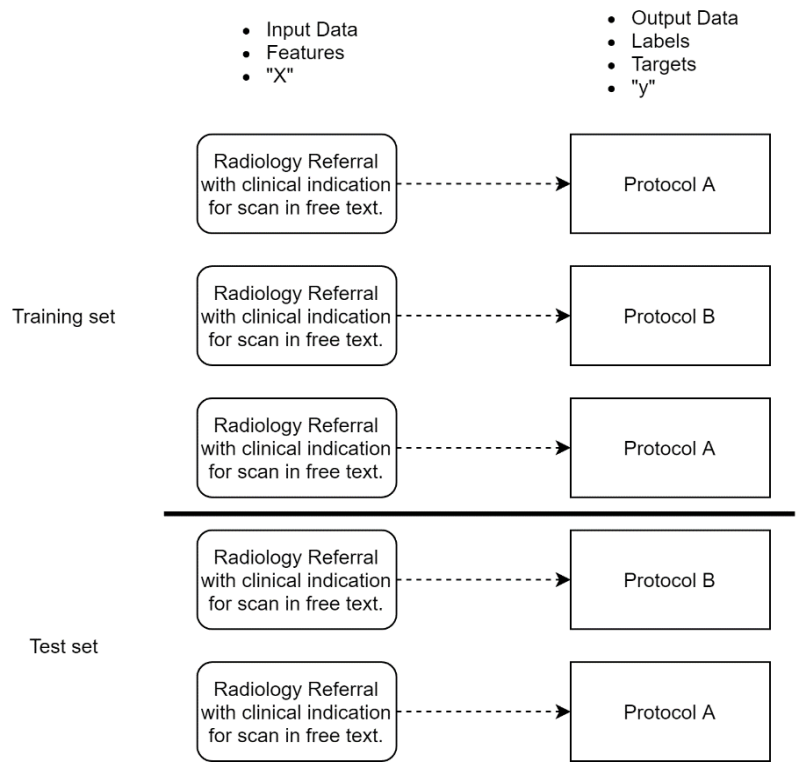

*Figure 3: An example of a simple ML text classification dataset with 5 samples (3 in the training set, 2 in the test set). See Internal Data section for discussion of training and test sets.*

# Appendix i

Search Strategy

1. Radiology.mp. or exp Radiology/
2. exp Tomography, X-Ray Computed/ or Computed Tomography.mp.
3. exp Magnetic Resonance Imaging/ or MRI.mp.
4. 1 or 2 or 3
5. exp Automation/ or Automation.mp.
6. Natural Language Processing.mp. or exp Natural Language Processing/
7. Machine Learning.mp. or exp Natural Language Processing/
8. Protocol.mp. or Protocoling.mp. or Protocols.mp.
9. 5 or 6 or 7
10. 4 and 8 and 9
11. Limit 4 to (yr="2017 -Current" and english)

# Appendix ii

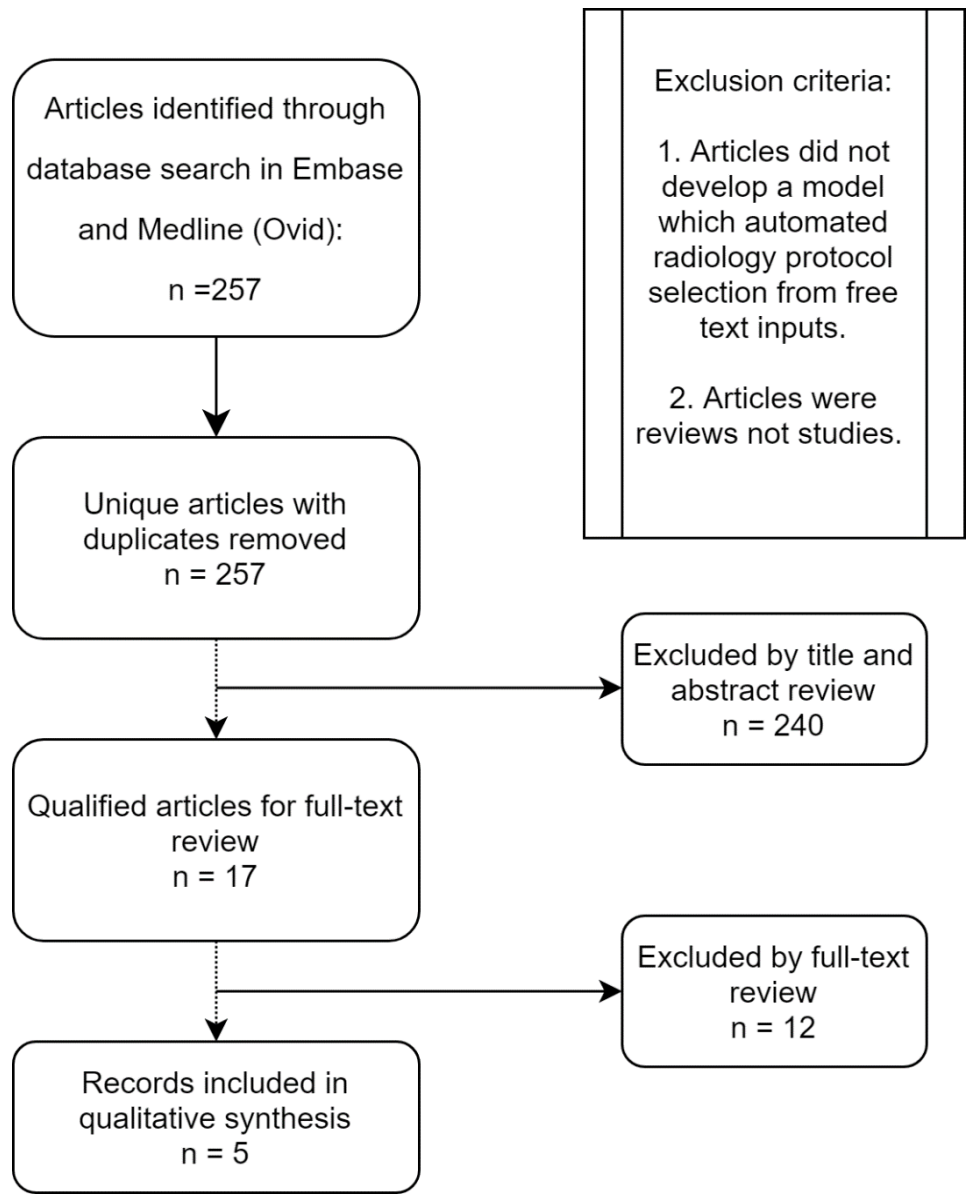